\documentclass[11pt]{article}
\usepackage[preprint]{acl}
\usepackage{times}
\usepackage{latexsym}
\usepackage{booktabs}
\usepackage{multirow}
\usepackage{amsmath} 
\usepackage{amssymb}
\usepackage[T1]{fontenc}
\usepackage[utf8]{inputenc}
\usepackage{microtype}
\usepackage{inconsolata}
\usepackage{graphicx}
\usepackage{xcolor}
\usepackage{array}

\title{Efficient Fine-Tuning Methods for Portuguese Question Answering: 
A Comparative Study of PEFT on BERTimbau and Exploratory Evaluation of Generative LLMs}

\author{Mariela M. Nina \and Caio Veloso Costa \and Lilian Berton \and Didier A. Vega-Oliveros \\
 \textit{Institute of Science and Technology}\\
\textit{Federal University of São Paulo (UNIFESP)}\\
São José dos Campos, SP, Brazil \\
  \small{
    \textbf{Correspondence:}
    \{\href{mailto:mariela.nina@unifesp.br}{mariela.nina}, 
    \href{mailto:veloso.caio@unifesp.br}{veloso.caio},
    \href{mailto:lberton@unifesp.br}{lberton},
    \href{mailto:didier.vega@unifesp.br}{didier.vega}\}@unifesp.br
  }}

\begin{document}
\maketitle
\begin{abstract}
Although large language models have transformed natural language processing, their computational costs create accessibility barriers for low-resource languages such as Brazilian Portuguese. This work presents a systematic evaluation of Parameter-Efficient Fine-Tuning (PEFT) and quantization techniques applied to BERTimbau for Question Answering on SQuAD-BR, the Brazilian Portuguese translation of SQuAD v1.
We evaluate 40 configurations combining four PEFT methods (LoRA, DoRA, QLoRA, QDoRA) across two model sizes (Base: 110M, Large: 335M parameters). Our findings reveal three critical insights: (1) LoRA achieves 95.8\% of baseline performance on BERTimbau-Large while reducing training time by 73.5\% (F1=81.32 vs 84.86); (2) higher learning rates (2e-4) substantially improve PEFT performance, with F1 gains of up to +19.71 points over standard rates; and (3) larger models show twice the quantization resilience (loss of 4.83 vs 9.56 F1 points).
These results demonstrate that encoder-based models can be efficiently fine-tuned for extractive Brazilian Portuguese QA with substantially lower computational cost than large generative LLMs, promoting more sustainable approaches aligned with \textit{Green AI} principles. An exploratory evaluation of Tucano and Sabiá on the same extractive QA benchmark shows that while generative models can reach competitive F1 scores with LoRA fine-tuning, they require up to 4.2$\times$ more GPU memory and 3$\times$ more training time than BERTimbau-Base, reinforcing the efficiency advantage of smaller encoder-based architectures for this task.

\end{abstract}

\section{Introduction}
Large language models (LLMs) based on the Transformer architecture \cite{vaswani2017attention} have achieved extraordinary capabilities in recent years, reaching state-of-the-art results across multiple natural language processing benchmarks \cite{devlin2019bert}. However, in lower-resource language settings such as Brazilian Portuguese, the landscape faces significant limitations. Unlike English, which benefits from a proliferation of specialized models, Brazilian Portuguese has a more restricted ecosystem. The most relevant and widely used models, like Sabiá (7B parameters) \cite{pires2023sabia}, Tucano (1.1B parameters) \cite{correa2025tucano}, and BERTimbau (110M--335M parameters) \cite{souza2020bertimbau}, are primarily based on widely established architectures subsequently adapted for Portuguese. Although these models have demonstrated competitive performance, they are typically adapted using full parameter updates, which can demand significant computational resources (e.g., up to 7 hours and over 18\,GB of GPU memory for Large models, as observed in our experiments), resulting in accessibility barriers for academic and industrial environments with constrained resources \cite{strubell2019energy}.

To overcome these computational constraints, QLoRA \cite{dettmers2023qlora} has emerged as a promising technique. Unlike standard post-training quantization used solely for model compression during inference, QLoRA enables fine-tuning of 4-bit quantized models with low-rank adapters while keeping the base model frozen, without significant performance degradation. While this technique has been widely adopted in English-language models, studies applying quantization techniques to Brazilian Portuguese question-answering models remain scarce and largely unsystematic. Parameter-Efficient Fine-Tuning (PEFT) methods provide an additional solution. Techniques such as LoRA \cite{hu2021lora}, which injects low-rank matrices while updating only 0.1--1\% of the parameters, and DoRA \cite{liu2024dora}, which introduces magnitude--direction decomposition, have been shown to achieve performance close to full fine-tuning in English. Nevertheless, empirical evidence remains overwhelmingly concentrated on English-language models, leaving unanswered whether these techniques retain their effectiveness when applied to Brazilian Portuguese models for tasks requiring deep language understanding.

Driven by the limitations of current hardware and the lack of Portuguese benchmarks, this study is guided by three core hypotheses: \textbf{(H1) Low-Resource Efficiency:} We hypothesize that PEFT methods can match full fine-tuning performance on Portuguese QA while significantly reducing computational overhead. \textbf{(H2) Scale Robustness:} Based on scaling laws, we hypothesize that larger models (Large) possess greater parametric redundancy, making them more resilient to aggressive 4-bit quantization than Base models. \textbf{(H3) Optimization Sensitivity:} We hypothesize that the low-rank constraints of PEFT require significantly higher learning rates than full fine-tuning to escape local minima during adaptation.

This work addresses these gaps by presenting a systematic evaluation of PEFT and quantization techniques applied to Brazilian Portuguese models on the Portuguese version of the SQuAD v1 dataset to test these hypotheses. Focusing on BERTimbau, the most established Transformer-based model for Brazilian Portuguese, with well-defined baselines (F1 = 82.50 for Base and F1 = 84.43 for Large on Portuguese SQuAD v1) \cite{souza2020bertimbau}, we conduct a comparative evaluation of LoRA, QLoRA, DoRA, and QDoRA on both the Base (110M parameters) and Large (335M parameters) variants for the Question Answering task. We provide a comprehensive analysis of the trade-offs among performance (F1-score and Exact Match), temporal efficiency, and hardware accessibility, using full fine-tuning as the baseline.

\section{Background}

\subsection{The Challenge of Adapting Large Models}

The dominant paradigm in NLP consists of pre-training massive models on large corpora and adapting them via full fine-tuning for specific tasks \cite{devlin2019bert}. However, as models scale according to established scaling laws \cite{kaplan2020scaling,hoffmann2022training}, full fine-tuning becomes prohibitively expensive: it requires storing gradients and optimizer states for all parameters, demanding up to 12--18$\times$ more memory than inference. For example, fine-tuning GPT-3 175B with the Adam optimizer requires approximately 1.2TB of GPU memory \cite{hu2021lora}, making the deployment of multiple specialized model instances impractical. This resource barrier motivated the development of Parameter-Efficient Fine-Tuning (PEFT) methods, which aim to update only a small fraction of parameters while maintaining competitive performance \cite{hu2021lora,peft}.

\subsection{LoRA: Low-Rank Adaptation}

LoRA \cite{hu2021lora} addresses this challenge by building on two key observations: (1) pre-trained models exhibit low intrinsic dimensionality \cite{aghajanyan2020intrinsic}, suggesting that the effective adaptation space is much smaller than the full parameter space, and (2) weight updates during fine-tuning exhibit low-rank structure. Motivated by these findings, LoRA keeps the pre-trained weights $W_0 \in \mathbb{R}^{d \times k}$ frozen (where $d$ is the output dimension and $k$ is the input dimension) and injects two trainable low-rank matrices: $A \in \mathbb{R}^{r \times k}$ and $B \in \mathbb{R}^{d \times r}$, where the rank $r \ll \min(d, k)$ is typically chosen from $r \in \{4, 8, 16\}$ in practice.\footnote{We adopt the notation $d, k, r$ to maintain consistency with the original LoRA formulation \cite{hu2021lora}.} For an input $x \in \mathbb{R}^{k}$, the output $h \in \mathbb{R}^{d}$ is computed as:

\begin{equation}
h = W_0 x + \frac{\alpha}{r} B A x
\end{equation}

where $\alpha$ is a scaling factor, typically set to $\alpha = 2r$ to stabilize training. This decomposition drastically reduces the number of trainable parameters: for a $768 \times 768$ matrix with $r=16$, LoRA requires only $2 \times 16 \times 768 = 24{,}576$ parameters versus $768^2 = 589{,}824$ in full fine-tuning (a 96\% reduction). Crucially, during inference, $BA$ can be merged into $W_0$, eliminating any additional latency—an important advantage over adapter-based methods \cite{houlsby2019adapter}.

Other PEFT methods include Prefix-Tuning \cite{li2021prefix}, which optimizes task-specific continuous vectors prepended to input representations; Prompt Tuning \cite{lester2021power}, which learns soft prompts prepended to the input; and AdaLoRA \cite{zhang2023adalora}, which adaptively allocates the parameter budget based on the importance of each weight matrix. Recent studies propose a unified view of these methods \cite{he2022towards}, identifying common patterns in how they modify the base model's representations.

\subsection{Quantization and QLoRA}

Complementary to PEFT methods, quantization approaches efficiency from an orthogonal perspective: reducing the numerical precision of model weights from floating-point representations (typically 32 or 16 bits) to lower-precision formats (8, 4, 3, or 2 bits) \cite{dettmers2023qlora}. For uniform $n$-bit quantization, a weight $w \in \mathbb{R}$ is mapped to a quantized integer $\tilde{w}$ as:

\begin{equation}
\tilde{w} = \text{round}\left(\text{clamp}\left(\frac{w - z}{s}, -2^{n-1}, 2^{n-1}-1\right)\right)
\end{equation}

where $s \in \mathbb{R}^+$ is the scale factor and $z \in \mathbb{R}$ is the zero-point. Prior work explores 8-bit optimizers \cite{dettmers20218bit} to reduce memory footprint during training, while GPTQ \cite{frantar2023gptq} proposes post-training quantization based on layer-wise error minimization.

QLoRA \cite{dettmers2023qlora} represents a synergistic integration of quantization and LoRA, for which we show evidence here that 4-bit-quantized models can be fine-tuned without performance degradation. QLoRA introduces three key technical innovations: (1) \textbf{4-bit NormalFloat (NF4)}, a data type designed for normally distributed weights that uses quantile-based quantization and is theoretically optimal for deep neural networks; (2) \textbf{Double Quantization}, which also quantizes the scale and zero-point parameters, reducing the average footprint by approximately 0.37 bits per parameter; and (3) \textbf{Paged Optimizers}, which leverage unified memory to manage memory spikes. In QLoRA, the base weights are quantized to 4 bits and remain frozen, while the LoRA matrices are kept in bfloat16.

\subsection{DoRA and QDoRA}

DoRA \cite{liu2024dora} addresses the persistent accuracy gap of LoRA by decomposing weights into magnitude and direction components. Inspired by Weight Normalization \cite{salimans2016weight}, DoRA decomposes each weight matrix $W$ as:

\begin{equation}
W = m \frac{V}{\|V\|_c}
\end{equation}

where $m \in \mathbb{R}^{d}$ represents the magnitudes (column-wise norms) and $V \in \mathbb{R}^{d \times k}$ represents the normalized direction. During fine-tuning, DoRA applies the LoRA update only to the directional component and trains an additional vector $\Delta m$ to adjust the magnitudes. QDoRA naturally extends DoRA to the quantized regime by combining this decomposition with QLoRA's quantization techniques.

\subsection{Question Answering and Metrics}

In extractive Question Answering (QA) tasks such as SQuAD v1 \cite{rajpurkar2016squad}, given a context $C$ and a question $Q$, the model must predict the start $s$ and end $e$ positions that delimit the answer span extracted from the context. The standard evaluation metrics are the \textbf{F1-score} (the harmonic mean of token-level precision and recall) and \textbf{Exact Match (EM)} (the percentage of predictions that exactly match the ground truth after normalization).

Beyond standalone applications, robust QA models serve as critical reasoning components in downstream pipelines, such as explainable automated fact-checking~\cite{Yang2022}, where QA mechanisms act as proxies for claims against retrieved evidence and provide interpretability.

\section{Related Work}

Early efforts on pre-trained models for Brazilian Portuguese focused on BERT-style architectures, with particular emphasis on BERTimbau in its Base and Large variants \cite{souza2020bertimbau}. Subsequent studies explored autoregressive LLMs such as Sabiá \cite{pires2023sabia} and Tucano \cite{correa2025tucano}, which were designed primarily for text generation. More recently, updated models such as Bertugues \cite{bertugues} have further expanded the landscape of Portuguese representation learning. While multilingual models such as XLM-R \cite{conneau2020unsupervised} demonstrate strong cross-lingual capabilities, BERTimbau's specific focus on Brazilian Portuguese makes it especially suitable for downstream tasks in this language.

In extractive QA, SQuAD-BR has become the standard benchmark for evaluating models in Portuguese, with BERTimbau Base and Large achieving reference results in F1 and Exact Match on the Portuguese version of SQuAD v1 (e.g., $F1 = 82.50\%$ and $84.43\%$, $EM = 70.49\%$ and $72.68\%$, respectively) \cite{souza2020bertimbau,kdmile}. While some recent studies have explored self-supervised fine-tuning and layer-freezing strategies to adapt BERTimbau for specialized domains with limited labeled data~\cite{Nina2025,Condori-Luna2026}, these works often prioritize classification tasks and do not explicitly analyze the computational costs of full fine-tuning in extractive QA.

In the context of English-language LLMs, Parameter-Efficient Fine-Tuning (PEFT) methods such as LoRA \cite{hu2021lora} and DoRA \cite{liu2024dora} have been proposed, introducing low-rank adaptations that drastically reduce the number of updated parameters while maintaining competitive performance. In parallel, quantization techniques such as QLoRA \cite{dettmers2023qlora} enable storing model weights at low precision (4 bits) and combining this with PEFT to train LLMs on GPUs with limited memory. While benchmarks such as GLUE \cite{wang2019glue} and SuperGLUE \cite{wang2019superglue} have established robust evaluation standards, they remain heavily English-centric, highlighting a significant representational gap for low-resource languages. For instance, techniques like progressive layer unfreezing have proven crucial for successfully optimizing large multilingual models in extremely low-resource settings, such as those required for indigenous languages~\cite{Nina2025}.

Overall, the existing literature provides: (i) robust Portuguese models with a focus on absolute performance, (ii) strong QA baselines on SQuAD-BR based on full fine-tuning, and (iii) a mature body of PEFT techniques evaluated primarily in English. However, systematic evidence that combines these three lines, i.e., evaluating PEFT and quantization on Brazilian Portuguese models for QA, remains scarce. This gap is precisely what the present work aims to address.

\section{Methodology}

\subsection{Dataset and Model Configuration}

We use SQuAD-BR \cite{rajpurkar2016squad}, consisting of 87,599 question–answer pairs for training and 10,570 for evaluation. To evaluate scale robustness (H2), we utilize two variants of BERTimbau \cite{souza2020bertimbau}: (1) \textbf{BERTimbau-Base} with 12 layers, 768 hidden dimensions, 12 attention heads, and 110M parameters, and (2) \textbf{BERTimbau-Large} with 24 layers, 1024 hidden dimensions, 16 attention heads, and 335M parameters. Both models were pre-trained on the brWaC corpus with 2.68 billion tokens.

\subsection{PEFT Configuration}

For all PEFT methods, we use: LoRA rank $r=16$, scaling factor $\alpha=32$, target modules (the query, key, value, and output projection matrices of the attention mechanism), and a dropout rate of 0.1. For quantized variants, we apply 4-bit NF4 quantization to the base weights with double quantization enabled and bfloat16 as the compute dtype. All prompts used for generative model evaluation are publicly available to ensure replicability.\footnote{\url{https://github.com/GPAM-ai/Efficient-FineTunning-QA-PEFT.git}}

To test our hypothesis on optimization sensitivity (H3), we systematically compare two learning rates: the standard BERT learning rate ($4.25\times 10^{-5}$) and a high learning rate optimized for PEFT ($2\times 10^{-4}$), training for 2 and 3 epochs. Common hyperparameters include the AdamW optimizer \cite{kingma2015adam,loshchilov2019decoupled}, weight decay of 0.01, batch size of 16 (Base) and 8 (Large), maximum sequence length of 384, and gradient clipping with norm 1.0.

\subsection{Computational Infrastructure}

All experiments were conducted on a workstation with a single GPU: NVIDIA RTX A4500 with 20GB of VRAM. This setup simulates a constrained academic environment (relevant to H1), where full fine-tuning of large models is typically unfeasible without techniques like QLoRA. Software stack: CUDA 12.2, PyTorch 2.1.0, Transformers 4.36.0, PEFT 0.7.1, and bitsandbytes 0.41.0.

\section{Experimental Results}

\subsection{Performance on BERTimbau-Base}

Tables \ref{tab:base_lr_high} and \ref{tab:base_lr_standard} present the complete results for BERTimbau-Base, considering variations in the \textit{learning rate} ($2\times10^{-4}$ and $4.25\times10^{-5}$) and the number of epochs (2 and 3), using full fine-tuning (\textit{Full FT}) solely as an upper reference for performance. The Full FT baseline values reported here are the result of our own re-execution under identical hardware and software conditions, enabling a fair comparison; they are close to, though not identical to, the original values reported by \citet{souza2020bertimbau}, as we expect minor differences due to distinct software versions and random seeds.

With $lr=2\times10^{-4}$, PEFT methods exhibit consistent and stable behavior, with \textbf{LoRA} and \textbf{DoRA} standing out as the best-performing techniques (F1=78.01), closely approaching the \textit{Full FT} baseline. From a practical perspective, LoRA is preferable because it achieves the same accuracy while reducing training time by 68.6\% and peak GPU memory by 74.6\% relative to full fine-tuning (3{,}687\,MB vs.\ 14{,}493\,MB; see Table~\ref{tab:memory}).

Figure~\ref{fig:heatmap} summarizes these trends across all configurations, showing the F1 scores for all methods and learning rate combinations.

\begin{table}[h]
\centering
\caption{BERTimbau-Base on SQuAD-BR (QA) with high learning rate ($2\times 10^{-4}$). Metrics: F1 and Exact Match (EM).}
\label{tab:base_lr_high}
\small
\begin{tabular}{lcccc}
\toprule
\textbf{Method} & \textbf{Ep.} & \textbf{F1} & \textbf{EM} & \textbf{Time} \\
\midrule
\textcolor{gray}{\textbf{Full FT}} & 2 & \textcolor{gray}{\textbf{79.74}} & \textcolor{gray}{\textbf{67.15}} & 01:40:02 \\
\textbf{LoRA} & 2 & \textbf{78.01} & \textbf{64.85} & 00:31:37 \\
QLoRA & 2 & 73.23 & 60.26 & 00:30:03 \\
\textbf{DoRA} & 2 & \textbf{78.01} & \textbf{64.89} & 00:40:23 \\
QDoRA & 2 & 74.41 & 61.26 & 00:42:03 \\
\midrule
\textcolor{gray}{\textbf{Full FT}} & 3 & \textcolor{gray}{\textbf{78.33}} & \textcolor{gray}{\textbf{65.54}} & 02:29:04 \\
\textbf{LoRA} & 3 & \textbf{78.01} & \textbf{65.03} & 00:46:47 \\
QLoRA & 3 & 74.16 & 61.24 & 00:44:42 \\
\textbf{DoRA} & 3 & \textbf{78.27} & \textbf{65.08} & 00:59:59 \\
QDoRA & 3 & 74.46 & 61.32 & 01:02:44 \\
\bottomrule
\end{tabular}
\end{table}

\begin{table}[h]
\centering
\caption{BERTimbau-Base on SQuAD-BR (QA) with standard learning rate ($4.25\times 10^{-5}$). Metrics: F1 and Exact Match (EM).}
\label{tab:base_lr_standard}
\small
\begin{tabular}{lcccc}
\toprule
\textbf{Method} & \textbf{Ep.} & \textbf{F1} & \textbf{EM} & \textbf{Time} \\
\midrule
\textcolor{gray}{\textbf{Full FT}} & 2 & \textcolor{gray}{\textbf{82.79}} & \textcolor{gray}{\textbf{70.91}} & 01:40:04 \\
\textbf{LoRA} & 2 & \textbf{71.81} & \textbf{58.07} & 00:31:49 \\
QLoRA & 2 & 53.52 & 40.54 & 00:30:02 \\
DoRA & 2 & 71.36 & 57.68 & 00:40:13 \\
QDoRA & 2 & 54.10 & 41.15 & 00:42:10 \\
\midrule
\textcolor{gray}{\textbf{Full FT}} & 3 & \textcolor{gray}{\textbf{82.18}} & \textcolor{gray}{\textbf{70.40}} & 02:28:52 \\
\textbf{LoRA} & 3 & \textbf{72.01} & \textbf{58.32} & 00:41:30 \\
QLoRA & 3 & 53.19 & 39.81 & 00:40:20 \\
DoRA & 3 & 71.50 & 57.65 & 00:53:58 \\
QDoRA & 3 & 58.42 & 45.37 & 00:55:00 \\
\bottomrule
\end{tabular}
\end{table}

In contrast, with the standard \textit{learning rate} ($lr=4.25\times10^{-5}$), the performance of PEFT methods degrades significantly. LoRA reaches only F1=71.81 (86.7\% of \textit{Full FT}), while the quantized variants almost completely collapse (QLoRA: F1=53.52, QDoRA: F1=54.10), losing more than 20 F1 points relative to full fine-tuning. Using $lr=2\times10^{-4}$, LoRA improves by \textbf{+6.20 F1 points} (F1=78.01, 94.2\% of \textit{Full FT}) and QLoRA recovers \textbf{+19.71 F1 points} (F1=73.23), while reducing peak memory to just 1{,}897\,MB (86.9\% reduction vs.\ full fine-tuning).

\subsection{Performance on BERTimbau-Large}

Tables \ref{tab:large_lr_high} and \ref{tab:large_lr_standard} present the complete results for BERTimbau-Large. At a high learning rate ($lr=2\times10^{-4}$), the Full FT baseline collapses critically (F1=3.02 and F1=5.14 at 2 and 3 epochs, respectively), while PEFT methods maintain robust performance. In particular, {LoRA} reaches F1=81.32 (95.8\% of the optimal \textit{Full FT} baseline) with a 73.5\% reduction in training time and 50.2\% reduction in peak memory (9{,}019\,MB vs.\ 18{,}125\,MB). QLoRA achieves F1=80.03 while requiring only 3{,}281\,MB—an 81.9\% memory reduction—demonstrating the feasibility of training large models under severe hardware constraints (Table~\ref{tab:memory}).

\begin{figure*}[ht!]
\centering
\includegraphics[width=\textwidth]{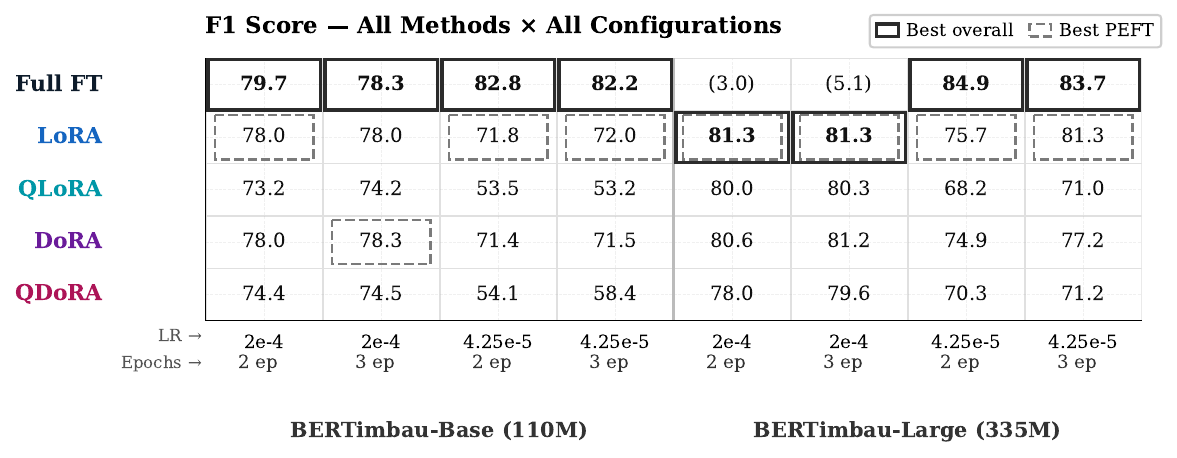}
\caption{F1 score for all PEFT methods across all configurations (learning rate $\times$ epochs $\times$ architecture). A {bold border} remarks the single best overall result per column; {gray dashed borders} show the best PEFT method per column. When both criteria coincide (i.e., a PEFT method is also the best overall), they are shown together. Values in parentheses indicate training collapse of full fine-tuning (\emph{Full FT}) on BERTimbau-Large under high learning rate (lr=$2\times10^{-4}$), observed across both 2 and 3 epoch settings.}
\label{fig:heatmap}
\end{figure*}

Figure~\ref{fig:heatmap} further illustrates this behavior: while Full FT collapses at high learning rates (F1=3.02 and F1=5.14, shown in parentheses for 2 and 3 epochs), PEFT methods remain stable, suggesting that the low-rank structure of LoRA and DoRA acts as an implicit regularizer, preventing divergence during training.

\begin{table}[h]
\centering
\caption{BERTimbau-Large on SQuAD-BR (QA) with high learning rate ($2\times 10^{-4}$). Metrics: F1 and Exact Match (EM).}
\label{tab:large_lr_high}
\small
\begin{tabular}{lcccc}
\toprule
\textbf{Method} & \textbf{Ep.} & \textbf{F1} & \textbf{EM} & \textbf{Time} \\
\midrule
\textcolor{gray}{\textbf{Full FT}} & 2 & \textcolor{gray}{\textbf{3.02}} & \textcolor{gray}{\textbf{0.03}} & 05:15:30 \\
\textbf{LoRA} & 2 & \textbf{81.32} & \textbf{68.67} & 01:23:41 \\
QLoRA & 2 & 80.03 & 67.17 & 01:19:15 \\
DoRA & 2 & 80.61 & 68.09 & 01:47:37 \\
QDoRA & 2 & 77.96 & 65.05 & 01:57:30 \\
\midrule
\textcolor{gray}{\textbf{Full FT}} & 3 & \textcolor{gray}{\textbf{5.14}} & \textcolor{gray}{\textbf{0.11}} & 07:50:02 \\
\textbf{LoRA} & 3 & \textbf{81.27} & \textbf{68.67} & 02:05:20 \\
QLoRA & 3 & 80.28 & 67.63 & 01:57:39 \\
DoRA & 3 & 81.22 & 68.70 & 02:40:52 \\
QDoRA & 3 & 79.61 & 66.99 & 02:54:52 \\
\bottomrule
\end{tabular}
\end{table}

\begin{table}[h]
\centering
\caption{BERTimbau-Large on SQuAD-BR (QA) with standard learning rate ($4.25\times 10^{-5}$). Metrics: F1 and Exact Match (EM).}
\label{tab:large_lr_standard}
\small
\begin{tabular}{lcccc}
\toprule
\textbf{Method} & \textbf{Ep.} & \textbf{F1} & \textbf{EM} & \textbf{Time} \\
\midrule
\textcolor{gray}{\textbf{Full FT}} & 2 & \textcolor{gray}{\textbf{84.86}} & \textcolor{gray}{\textbf{73.00}} & 05:15:39 \\
\textbf{LoRA} & 2 & \textbf{75.65} & \textbf{62.21} & 01:23:28 \\
QLoRA & 2 & 68.23 & 54.92 & 01:19:12 \\
DoRA & 2 & 74.93 & 62.02 & 01:47:46 \\
QDoRA & 2 & 70.32 & 56.88 & 01:57:30 \\
\midrule
\textcolor{gray}{\textbf{Full FT}} & 3 & \textcolor{gray}{\textbf{83.74}} & \textcolor{gray}{\textbf{72.04}} & 07:50:46 \\
\textbf{LoRA} & 3 & \textbf{81.28} & \textbf{68.63} & 02:05:08 \\
QLoRA & 3 & 71.03 & 57.66 & 01:58:54 \\
DoRA & 3 & 77.18 & 63.98 & 02:41:23 \\
QDoRA & 3 & 71.24 & 58.15 & 02:55:23 \\
\bottomrule
\end{tabular}
\end{table}

With $lr=2\times10^{-4}$, LoRA achieves F1=81.32 (95.8\% of \textit{Full FT} baseline), whereas with $lr=4.25\times10^{-5}$ its performance drops to F1=75.65 (89.1\%), a difference of \textbf{+5.67 F1 points}. QLoRA drops from F1=80.03 to 68.23, losing 11.80 F1 points under the standard \textit{learning rate}. Crucially, QLoRA on Large shows a degradation of only $-$4.83 F1 points (vs.\ $-$9.56 on Base), an approximate $2\times$ difference that confirms greater quantization resilience in larger models \textbf{(H2)}.

An additional finding is the critical collapse of full fine-tuning with $lr = 2\times 10^{-4}$ (F1=3.02). This optimization divergence occurs because the high learning rate destabilizes the pre-trained representations across the full parameter space, whereas. In contrast, the structures of LoRA and DoRA act as implicit regularizers, preventing the model from drifting too far from the pre-trained manifold.

\subsection{Peak GPU Memory Consumption}

Table~\ref{tab:memory} consolidates peak GPU memory usage for both BERTimbau
variants and the generative models evaluated in Section~5.5, enabling a direct comparison of hardware requirements across all architectures.

\begin{table}[t]
\centering
\caption{Peak GPU memory (MB) during training on SQuAD-BR.}
\label{tab:memory}
\small
\begin{tabular}{lr}
\toprule
\textbf{Method / Setting} & \textbf{Peak Memory (MB)} \\
\midrule
\multicolumn{2}{l}{\textit{BERTimbau (Extractive QA, Training)}} \\
\cmidrule(lr){1-2}
Full Fine-tuning (Base)  & 14{,}493 \\
LoRA (Base)              &  3{,}687 \\
DoRA (Base)              &  4{,}295 \\
QLoRA (4-bit, Base)      &  1{,}897 \\
QDoRA (4-bit, Base)      &  2{,}163 \\
\cmidrule(lr){1-2}
Full Fine-tuning (Large) & 18{,}125 \\
LoRA (Large)             &  9{,}019 \\
DoRA (Large)             & 10{,}659 \\
QLoRA (4-bit, Large)     &  3{,}281 \\
QDoRA (4-bit, Large)     &  3{,}331 \\
\midrule
\multicolumn{2}{l}{\textit{Generative LLMs (LoRA Fine-tuning)}} \\
\cmidrule(lr){1-2}
Tucano-1B (Zero-shot)    &  2{,}682 \\
Tucano-1B + LoRA         &  4{,}301 \\
Sabiá-7B  (Zero-shot)    & 17{,}737 \\
Sabiá-7B  + LoRA         & 15{,}642 \\
\bottomrule
\end{tabular}
\end{table}

\begin{figure}[h]
\centering
\includegraphics[width=0.5\textwidth]{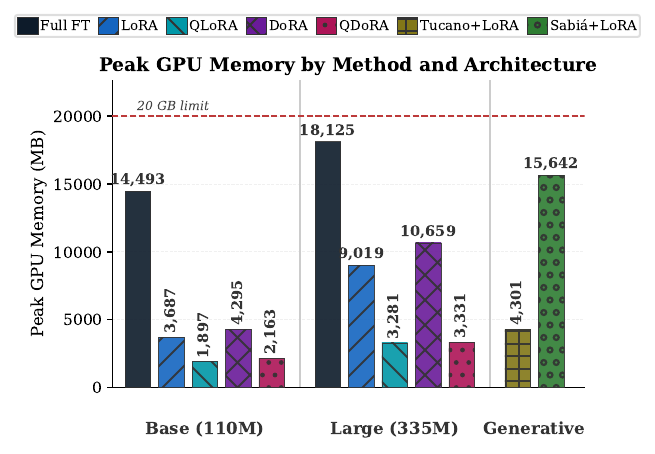}
\caption{Peak GPU memory consumption per method and architecture. Quantized methods (QLoRA, QDoRA) reduce memory by up to 86.9\% for Base and 81.9\% for Large relative to Full FT. Memory values for generative models (Tucano-1B, Sabiá-7B) are included for cross-architecture comparison; results for those models are discussed in Section~\ref{sec:generative}. The dashed red line marks the 20\,GB GPU limit of the hardware used.}
\label{fig:ram}
\end{figure}

Among BERTimbau variants, QLoRA achieves the largest reductions: 86.9\% for  ase (1{,}897\,MB vs.\ 14{,}493\,MB) and 81.9\% for Large (3{,}281\,MB vs.\ 18{,}125\,MB), both well within the range of consumer-grade GPUs. Notably, Sabiá-7B shows higher memory consumption at zero-shot (17{,}737\,MB) than with LoRA training (15{,}642\,MB), since zero-shot corresponds to full-precision inference while LoRA fine-tuning uses mixed-precision (4-bit base + bfloat16 adapters). Comparing across architectures, BERTimbau-Base with LoRA (3{,}687\,MB) achieves a similar F1 score to Sabiá-7B with LoRA (15{,}642\,MB) while requiring \textbf{4.2$\times$ less memory} and \textbf{3$\times$ less training time}; a compelling efficiency advantage for resource-constrained environments.

\subsection{Critical Impact of the Learning Rate}

The experiments confirm that the learning rate is the most decisive factor for the success of PEFT methods, consistent with our hypothesis on optimization sensitivity (H3), surpassing the influence of the specific method, the number of epochs, or the application of quantization. Figure~\ref{fig:heatmap} makes this effect directly visible: the left columns (lr=$2\times10^{-4}$) show consistently higher F1 than the right columns (lr=$4.25\times10^{-5}$) for all PEFT methods.

For both model sizes, higher learning rates ($lr=2\times10^{-4}$) are critical for maximizing performance. In BERTimbau-Base, LoRA improves performance by \textbf{+6.20 F1 points} relative to the standard \textit{learning rate}, whereas in BERTimbau-Large the gain is \textbf{+5.67 F1 points}. This pattern is further amplified under quantization: QLoRA exhibits gains of \textbf{+19.71 F1 points} in Base and \textbf{+11.80 F1 points} in Large when higher \textit{learning rates} are employed. Standard learning rates prove inadequate for PEFT schemes in our experiments, leading to severe degradation and, for Large models, complete \textit{Full FT} collapse.

\subsection{Exploratory Evaluation with Generative Models: Tucano and Sabiá}
\label{sec:generative}

To analyze the feasibility of autoregressive generative models for extractive Question Answering, we conducted an exploratory evaluation of Tucano \cite{correa2025tucano} and Sabiá \cite{pires2023sabia} on the same SQuAD-BR evaluation set used for BERTimbau. Unlike BERTimbau, these models are designed for free-text generation, which introduces a structural mismatch with SQuAD's exact-span extraction evaluation protocol. Consequently, we interpret these results as indicative of task-architecture compatibility rather than a direct performance comparison. Therefore, they should not be interpreted as a direct benchmark against encoder-based models. We evaluated two configurations: \textit{zero-shot} inference and fine-tuning with LoRA.

\begin{table}[h]
\centering
\caption{Exploratory results with generative models for extractive QA.}
\label{tab:llm_exploratory}
\footnotesize
\begin{tabular}{l l c c r}
\toprule
\textbf{Model} & \textbf{Config.} & \textbf{EM (\%)} & \textbf{F1 (\%)} & \textbf{Time} \\
\midrule
Tucano & Zero-shot &  4.02 & 14.68 & 00:16:59 \\
Sabiá  & Zero-shot & 25.37 & 39.82 & 00:55:26 \\
Tucano & LoRA      & 49.30 & 63.86 & 00:25:28 \\
Sabiá  & LoRA      & 64.11 & 78.10 & 01:31:15 \\
\midrule
\multicolumn{5}{l}{\textit{For reference (same evaluation set):}} \\
B-Base  & LoRA & 64.85 & 78.01 & 00:31:37 \\
B-Large & LoRA & 68.67 & 81.32 & 01:23:41 \\
\bottomrule
\end{tabular}
\end{table}

Table~\ref{tab:llm_exploratory} reveals that LoRA fine-tuning substantially improves both generative models: Tucano improves from F1=14.68\% to F1=63.86\% (+49.18 points), while Sabiá improves from F1=39.82\% to F1=78.10\% (+38.28 points). Notably, Sabiá with LoRA (F1=78.10\%) reaches a  erformance level comparable to BERTimbau-Base with LoRA (F1=78.01\%), and BERTimbau-Large (F1=81.32\%) still achieves the best overall result. However, this competitive F1 comes at a substantial computational cost: Sabiá-7B with LoRA requires 15{,}642\,MB of GPU memory and 01:31:15 of training time, compared to 3{,}687\,MB and 00:31:37 for BERTimbau-Base—a \textbf{4.2$\times$ memory overhead} and \textbf{3$\times$ longer training} for an equivalent F1 score.
These results confirm that, for extractive QA in Brazilian Portuguese, encoder-based architectures offer a superior efficiency-performance trade-off, particularly in resource-constrained environments.

\section{Discussion}

The results show that PEFT techniques preserve most of the performance of full fine-tuning while achieving substantial cost reductions. BERTimbau-Large achieves 95.8\% of baseline performance (F1=81.32 vs 84.86) while reducing training time by 73.5\% and peak memory by 50.2\%. In BERTimbau-Base, LoRA retains 94.2\% of the baseline performance while achieving a 68.6\% reduction in training-time, reinforcing the viability of PEFT in resource-constrained scenarios. These results are close to the classical BERTimbau baselines reported for SQuAD-BR \cite{souza2020bertimbau,kdmile}, validating our experimental configuration and confirming the hypothesis on low-resource efficiency (H1).

The learning rate emerges as the most decisive factor for PEFT success, consistent with the hypothesis on optimization sensitivity (H3). With \(lr = 2\times10^{-4}\), LoRA and DoRA achieve their best performance, whereas with the standard learning rate (\(4.25\times10^{-5}\)), the performance degrades consistently, with losses of up to 6.20 F1 points in Base and 5.67 points in Large. This effect becomes critical under quantization: QLoRA improves by +19.71 F1 points in Base when the learning rate is increased, indicating that quantized PEFT schemes require different optimization dynamics than full fine-tuning.

A key finding is the opposite behavior of full fine-tuning under high learning rates. In BERTimbau-Large, full fine-tuning collapses almost completely (F1=3.02 with \(lr = 2\times10^{-4}\)), while PEFT methods remain stable. This contrast suggests that the low-rank updates in LoRA and DoRA serve as implicit regularizers, limiting the magnitude of parameter updates even with aggressive learning rates, thereby preventing optimization divergence.

Four-bit quantization introduces a size-dependent degradation. In BERTimbau-Base, QLoRA loses 9.56 F1 points relative to the optimal baseline, whereas in BERTimbau-Large this loss is reduced to 4.83 points, retaining 94.3\% of baseline performance. This approximate $2\times$ difference confirms \textbf{(H2)}. While QLoRA underperforms standard LoRA in F1-score, this slight degradation is heavily outweighed by its memory reduction: 86.9\% for Base (1{,}897\,MB) and 81.9\% for Large (3{,}281\,MB), well within the range of consumer-grade GPUs and effectively eliminating out-of-memory errors in resource-constrained environments.

The comparison between LoRA and DoRA shows that DoRA introduces a consistent temporal overhead of approximately 28\% in BERTimbau-Large without clear improvements in F1 or Exact Match. LoRA thus emerges as the more efficient and preferred option: it closely approaches the classical BERTimbau baselines \cite{souza2020bertimbau,kdmile}, significantly reduces training time and energy consumption, and, when combined with controlled quantization on Large models, enables near state-of-the-art performance under strict computational constraints.

The exploratory evaluation with Tucano and Sabiá highlights a critical efficiency gap between encoder and decoder architectures. While Sabiá-7B with LoRA achieves a competitive F1 score (78.10\%) comparable to BERTimbau-Base (78.01\%), this comes at the cost of 4.2$\times$ more GPU memory (15{,}642\,MB vs.\ 3{,}687\,MB) and 3$\times$ longer training time. BERTimbau-Large with LoRA (F1=81.32\%) surpasses all generative models while requiring significantly fewer computational resources. These findings confirm that for extractive QA in Brazilian Portuguese, smaller encoder-based architectures offer a clearly superior efficiency-performance trade-off. In fact, recent studies have demonstrated that Portuguese-native generative models excel when applied to well-aligned generative tasks, such as translating natural language to SQL~\cite{Freitas2025}, reinforcing that task-architecture alignment is a key factor in model selection.

\section{Conclusions}

This work presented a systematic evaluation of PEFT and quantization techniques applied to BERTimbau for Question Answering on SQuAD-BR. We demonstrated that it is possible to achieve competitive performance while substantially reducing computational cost. LoRA on BERTimbau-Large reaches 95.8\% of baseline performance while reducing training time by 73\%. In contrast, QLoRAoRA preserves 94.3\% of baseline performance while cutting peak GPU memory by 81.9\% (3{,}281\,MB), well within the range of consumer-grade GPUs.

Experimental results validate the three core hypotheses driving this study. First, regarding low-resource efficiency \textbf{(H1)}, PEFT methods drastically reduce computational and memory overhead while maintaining near-baseline performance. Second, concerning scale robustness \textbf{(H2)}, larger models show approximately $2\times$ better resilience to aggressive 4-bit quantization (4.83 vs 9.56 F1 loss), positioning BERTimbau-Large with QLoRA as the optimal choice for GPU memory-constrained scenarios. Third, validating optimization sensitivity \textbf{(H3)}, higher learning rates ($2\times 10^{-4}$) appear critical for PEFT success, substantially improving F1 by up to +19.71 points over standard rates in our experiments. Additionally, DoRA offers no practical advantages over LoRA, matching its performance at the cost of a 28\% increase in training time.

The exploratory evaluation confirms that while generative models such as Sabiá-7B can reach competitive F1 scores with LoRA fine-tuning (78.10\% vs.\ 78.01\% for BERTimbau-Base), they require 4.2$\times$ more GPU memory and 3$\times$ more training time. BERTimbau-Large with LoRA (F1=81.32\%) achieves the best overall result at a fraction of the computational cost, reinforcing that smaller encoder-based architectures offer a superior efficiency-performance trade-off for extractive QA in resource-constrained environments.

The findings provide concrete guidelines for sustainable QA research: prioritize BERTimbau-Large with LoRA or QLoRA under high learning rates when GPUs with limited VRAM are available; use full fine-tuning only as a reference baseline; and avoid aggressive quantization on Base models unless memory constraints are extreme. The 73.5\% reduction in training time makes PEFT techniques particularly valuable for resource-constrained environments, promoting more accessible NLP practices aligned with \textit{Green AI} principles \cite{strubell2019energy}.

As future work, we propose extending these models to broader Portuguese datasets and downstream applications such as explainable fact-checking~\cite{Yang2022}, while exploring automated hyperparameter searches that jointly optimize performance and computational cost. By integrating PEFT with adaptive strategies, like progressive layer unfreezing~\cite{Nina2025}, we also aim to extend these results to indigenous and underrepresented languages, thereby driving broader digital inclusion.

\section*{Acknowledgements}

This study was financed in part by the Coordenação de Aperfeiçoamento de Pessoal de Nível Superior – Brasil (CAPES) – Finance Code 001 and Conselho Nacional de Desenvolvimento Científico e Tecnológico (CNPq).

The authors also acknowledge the use of Anthropic's Claude Sonnet 4.6 and Grammarly for language editing and correction of English errors throughout the manuscript. All scientific content, experimental design, and interpretations are the sole responsibility of the authors.

\bibliography{references}

\end{document}